\documentclass[11pt]{article}

\usepackage[]{acl}

\usepackage{times}
\usepackage{latexsym}

\usepackage{adjustbox}
\usepackage{multirow}
\usepackage{stmaryrd}
\usepackage{booktabs}
\usepackage{amssymb}
\usepackage{amsmath}
\usepackage{subfigure}
\usepackage{amstext}
\usepackage{url}

\usepackage[T1]{fontenc}

\usepackage[utf8]{inputenc}

\usepackage{microtype}

\usepackage{colortbl}
\usepackage{makecell}
\def\best{\bf\cellcolor[gray]{0.85}}
\def\secbest{\cellcolor[gray]{0.92}}
\title{Continually Detection, Rapidly React: Unseen Rumors Detection based on Continual Prompt-Tuning}

\author{
	Yuhui Zuo$^{1}$, 
	Wei Zhu$^{2}$, 
	Guoyong Cai$^{1}$,  \\
	$^1$Guilin University of Electronic Technology, Guilin, China \\
	$^2$East China Normal University, Shanghai, China \\
	\texttt{zzuuoo2019ok@gmail.com,ccgycai@guet.edu.cn,wzhu@stu.ecnu.edu.cn} \\}

\begin{document}
\maketitle

	\begin{abstract}
		Since open social platforms allow for a large and continuous flow of unverified information, rumors can emerge unexpectedly and spread quickly. However, existing rumor detection (RD) models often assume the same training and testing distributions and can not cope with the continuously changing social network environment. 
		This paper proposed a Continual Prompt-Tuning RD (CPT-RD) framework, which avoids catastrophic forgetting (CF) of upstream tasks during sequential task learning and enables bidirectional knowledge transfer between domain tasks.
		Specifically, we propose the following strategies: (a) Our design explicitly decouples shared and domain-specific knowledge, thus reducing the interference among different domains during optimization; (b) Several technologies aim to transfer knowledge of upstream tasks to deal with emergencies; (c) A task-conditioned prompt-wise hypernetwork (TPHNet) is used to consolidate past domains.
		In addition,  CPT-RD avoids CF without the necessity of a rehearsal buffer.
		Finally, CPT-RD is evaluated on English and Chinese RD datasets and is effective and efficient compared to prior state-of-the-art methods.
	\end{abstract}
	
	\section{Introduction}
	\label{sec:intro}
	\begin{figure}[!t]
		\centering
		\includegraphics[width=8.1cm,height=6.9cm]{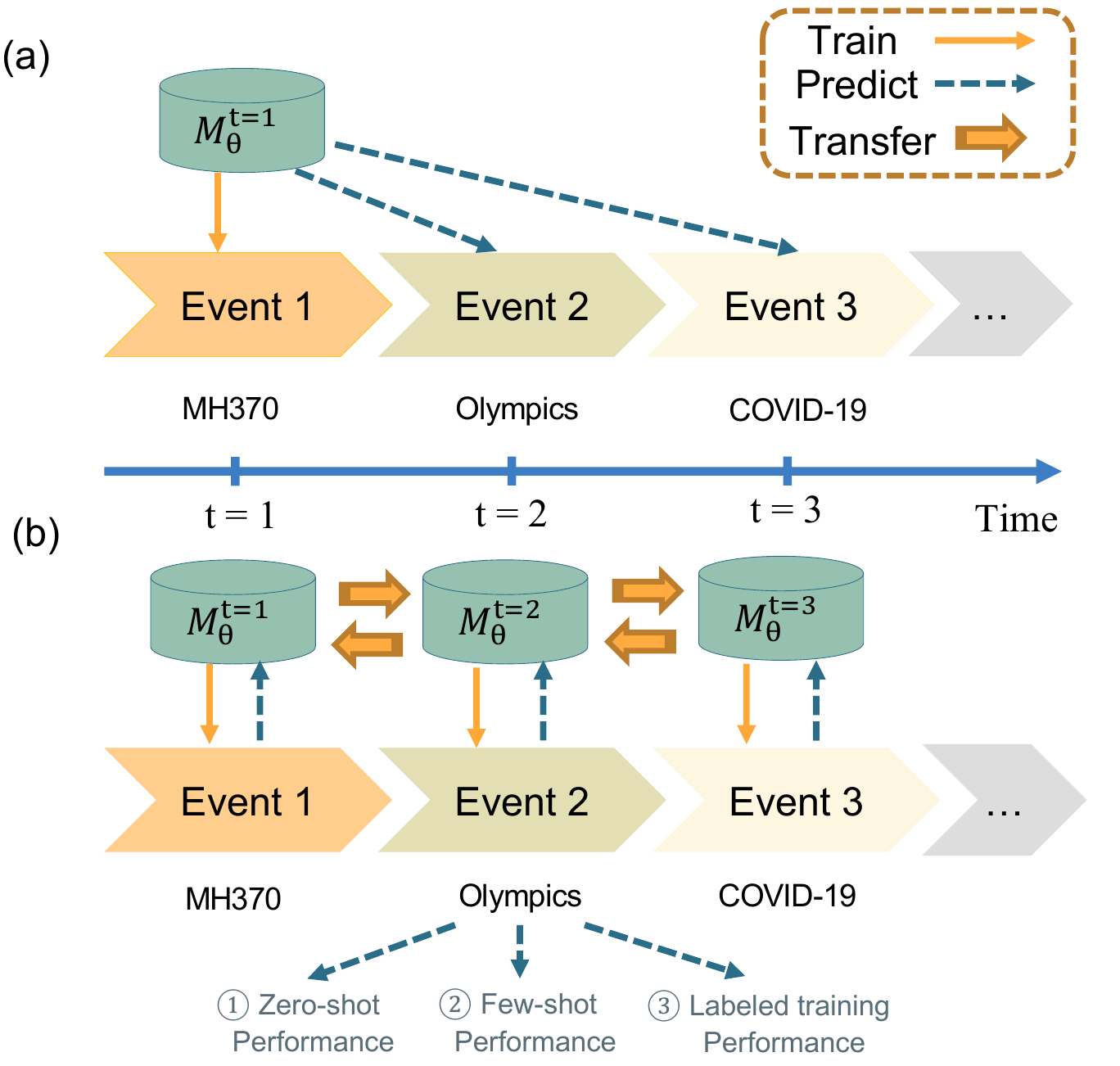}
		\caption{Illustration of the RD model training procedure in the traditional static (a) and continual dynamic event transfer (b) setup.}
		\label{fig:setup1}
		\vspace{-0.5cm}
	\end{figure}
	Online platforms such as social media are facing new and ever-evolving cyber threats at the information level --- rumor. 
	A rumor is an unconfirmed claim related to an object, event, or issue of public concern that is spread when its integrity is unknown \cite{guo2020future}. 
	It is very necessary to study automated RD, because rumors are extremely harmful to society and manual detection is time-consuming and labor-intensive \cite{oshikawa2018survey}.
	
	However, automated RD has significant challenges and still faces the following difficulties:
	First, rumors are highly event domain-specific \cite{wang2018eann}, each event domain may have a different input text distribution and fraudulent intent. 
	Second, detecting rumors at their early stage of spreading faces the problem of insufficient labeled samples \cite{zhou2020fake}.
	Third, rumor detectors operating on online social platforms often encounter continuous event domain changes.
	This poses a significant challenge to existing RD models.
	
	Previous work \cite{wang2018eann,bian2020rumor,zhang2021multi,lin2021rumor,ben2021pada,zhu2021gaml} usually assume the same distribution of training and testing data, and have difficulty coping with changing social network environments. 
	In Fig.\ref{fig:setup1} (a), there are no updates to the model regardless of how many unseen events will appear in the future, which requires the unreasonable assumption that the model's generalization capability is enough. 
	In addition, as the social network environment changes, the original training set data will become outdated \cite{lee2021dynamically}, so that the model's ability on more recent events will diminish. For example, the COVID-19 pandemic caused massive trouble for existing RD models \cite{lu2021novel,patwa2021fighting}.
	In social media, the ideal RD model must continuously detect event stream and react rapidly to emergencies.
	To study this ability, we propose a continual dynamic event transfer (CDET) setup (illustrated in Fig.\ref{fig:setup1} (b)), which makes the RD model dynamically updates the parameters $\theta$ over a series of sequential events and assume that each event goes through three stages from burst to normalization: zero-shot stage with no samples, few-shot stage with a small number of samples, and full-shot stage with large-scale labeled samples. This is because rumors must be detected early to avoid the social harm caused by their spread, but there are only a few or no labeled samples in the early stages of an event.
	
	Existing RD models still face many challenges in the CDET setup: (1) \textbf{Catastrophic forgetting:} When a neural model is trained in a sequence of tasks, the downstream tasks may catastrophically interfere with the upstream tasks. (2) \textbf{Knowledge transfer and accumulation:} Transfer the knowledge learned from upstream tasks for rapid generalization, accumulating knowledge from downstream tasks to better cope with upstream tasks; (3) \textbf{Parameter explosion:} Previous research \cite{wang2021mell,ke2021achieving} often require dynamically expanding neural modules for each task, which is undoubtedly aggravating for pre-trained language models (PLM) with billions of parameters, with limited memory; (4) \textbf{Data privacy:} After learning a task, training data is usually discarded due to user privacy concerns \cite{chen2018lifelong}. This requires models to share learned parameters, rather than saving data to retrain the model.
	
	
	
	

\begin{figure}[!t]
		\centering
		\includegraphics[width=7.3cm,height=6.5cm]{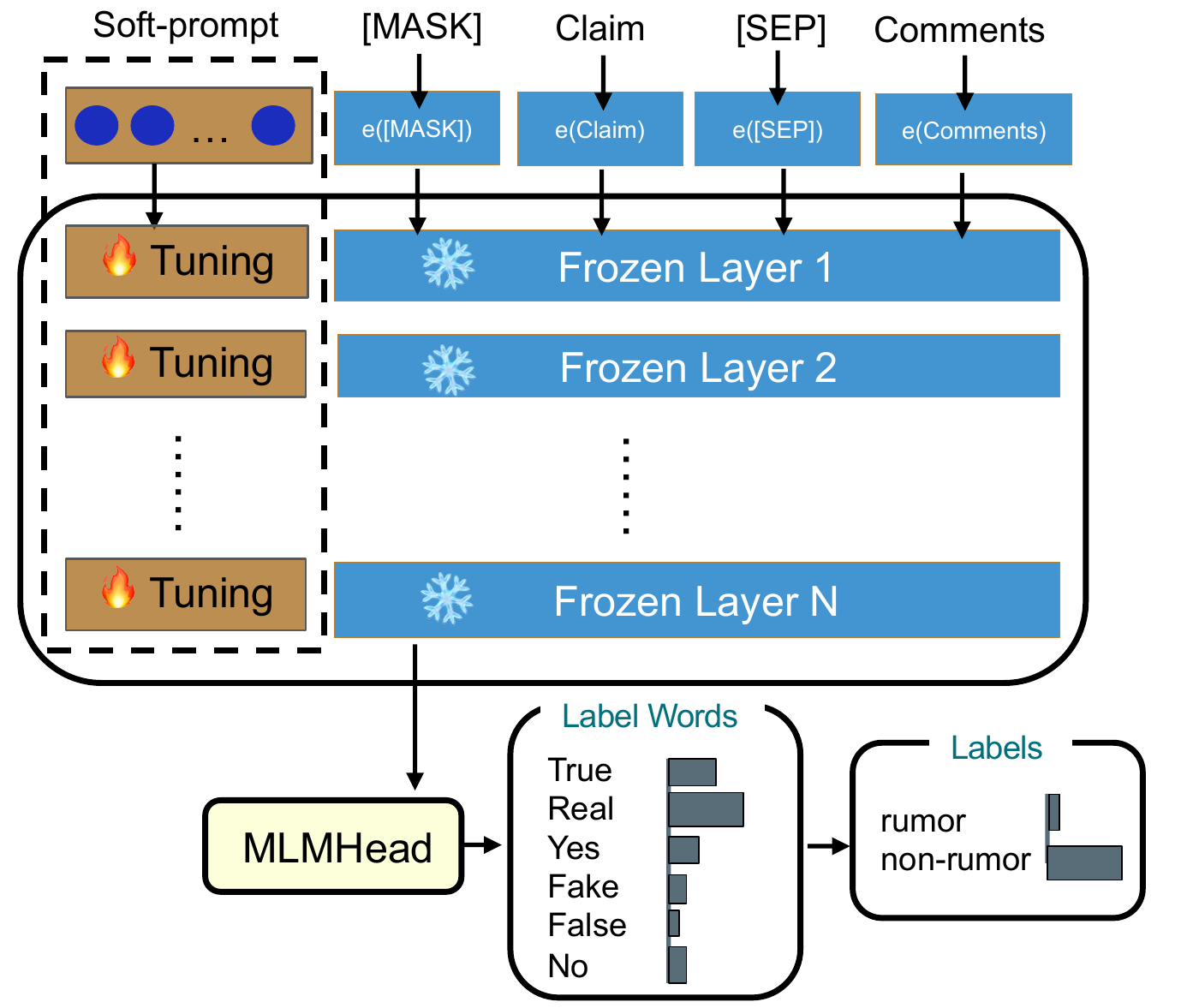}
		\caption{Illustration of deep prompt tuning for RD.}
		\label{fig:ptuning}
		\vspace{-0.5cm}
	\end{figure}

	To address the above challenges, a novel framework called \textbf{C}ontinual \textbf{P}rompt-\textbf{t}uning RD (CPT-RD) has been proposed. 
	Technically, CPT-RD can be seen as a continuously migrated version of P-tuning v2 \cite{liu2021p}.
	From Fig.\ref{fig:ptuning}, we can clearly decouple domain-specific knowledge (tuning parameter) and shared knowledge (frozen parameter).
	This provides the basis for achieving memory of task-specific parameters and bidirectional domain knowledge transfer.
	Bidirectional knowledge transfer includes:
	(1) \textbf{Forward Knowledge Transfer}: CPT-RD has various prompt initialization strategies to adapt fast to rumors of emergencies. (2) \textbf{Backward Knowledge Transfer}: A task-conditioned prompt-wise hypernetwork (TPHNet) learns latent distribution of soft-prompts, encouraging CPT-RD to accumulate knowledge in sequential events while avoiding data replay.
	
	

	We collected RD datasets with 14 different domains for both, English and Chinese. Through empirical analysis, we find that CPT-RD essentially avoids catastrophic forgetting. On the English dataset, the knowledge transfer indicators FWT and BWT \cite{lopez2017gradient} achieve positive indicators of 23.9\% and 0.9\%, respectively. Finally, the effectiveness of our improvement is demonstrated through ablation experiments.
	
	Our main contributions are: (1) We propose a CDET setup in RD for evaluating rapid generalization and continual detection problems simultaneously. 
	(2) To completely avoid the CF encountered in continual detection, we optimize and store domain-specific soft-prompt for each event domain and use it selectively. (3) We propose various forward knowledge transfer strategies to deal with early emergency rumors and accumulate detection experience through TPHNet for backward knowledge transfer. (4) Our experiments on the collected Chinese and English social media RD datasets demonstrate the superior performance and efficiency of our proposed method.
	
	
	\section{Related work}
	\label{sec:work}
	\subsection{Rumors detection}
	Social media has gained much attention as a source of research rumors, but existing RD methods still perform poorly in the face of unknown events and struggle to consistently respond to the dynamic and changing social network environment \cite{zubiaga2018detection,guo2020future}. \citet{wang2018eann} proposed an event adversarial neural network (EANN) which extracts event-invariant features by removing event-specific features. \citet{lu2021novel} consider unseen event RD as a few-shot learning problem, and \citet{wang2021multimodal} apply meta-learning to obtain the optimal generalization initial parameters. \citet{ben2021pada} used prompt-tuning to consider RD as domain adaptation and committed to improving out-of-distribution issues. The above methods can alleviate the problem of domain drift, but they are all learned with offline static settings. \citet{lee2021dynamically} considered the continual detection capability on online social media and applied rehearsal-based replay techniques to resist CF. However, this approach does not consider generalizing to the unseen domain quickly and faces problems such as inference efficiency and data privacy. 
	Different from the above methods, our proposed model considers both fast generalization and knowledge transfer for continual detection and is not based on any data replay.
	
	\subsection{Prompt-based tuning}
	Recent research has found that converting downstream tasks to language modeling tasks via textual prompts is more effective to use PLM than typical fine-tuning \cite{liu2021pre,sun2022simple,zhu-2021-leebert,zhu-2021-autorc}. Early prompting method, GPT-3 \cite{brown2020language} and PET/iPET \cite{schick2020s} for example, uses hand-crafted prompt templates.  However, the performance of these methods relies heavily on the selection of predefined prompt templates. Hand-crafting prompts are very time-consuming, and the performance may be sub-optimal. \citet{shin2020autoprompt} propose AutoPrompt to search for better prompts based on gradient descent approach. Instead of searching for discrete template words, \citet{li2021prefix} propose prefix-tuning, where tokens with trainable continuous embeddings are placed at the beginning of the text to perform generate tasks. P-tuning v2 \cite{liu2021p} also uses soft-prompt to achieve promising natural language understanding and knowledge probing tasks. Different from the above methods, they studied single-step adaptation, and we are interested in prompt transfer in CL environment.
	
	\subsection{Continual learning with fast generalization}
	Mitigating CF is usually a priority in CL or lifelong learning research proposals \cite{hadsell2020embracing}. 
	Recently, the demands on CL have increased further, not only to combat CF but also to generalize quickly in unseen tasks.
	Integration with meta-learning is a promising approach, concerned with balancing stability (preservation of past knowledge) and plasticity (rapid absorption of current knowledge).
	MER \cite{riemer2018learning} achieves gradient alignment by constraining the direction of the gradient angle between different task samples.
	Subsequently, OML \cite{javed2019meta} and La-MAML \cite{gupta2020maml} optimize and supplemente the training speed and effect of MER, respectively. Meta-MbPA \cite{wang2020efficient} combine the above meta-learning, rehearsal-based replay of CL and BERT.
	\citet{wang2021mell} uses global and local memory networks to capture different classes of cross-task prototype representations, adds a new frozen classification module for each task, and requires BERT to update slowly. \citet{ke2021adapting} and \citet{jin2021learn} freezes part of the backbone model against CF. They introduce an additional adapter layer \cite{houlsby2019parameter} to learn task-specific knowledge, avoiding inefficient data replay and reducing parameter tuning and growth rates. Different from the above methods, CPT-RD is based on PT. Although it is similar to adapter in terms of parameter tuning, PT is more effective in parameters, and CPT-RD has a more intuitive means of knowledge transfer and avoids data replay.
	

	\begin{figure*}[htp]
		\centering
		\includegraphics[width=15.3cm,height=8.7cm]{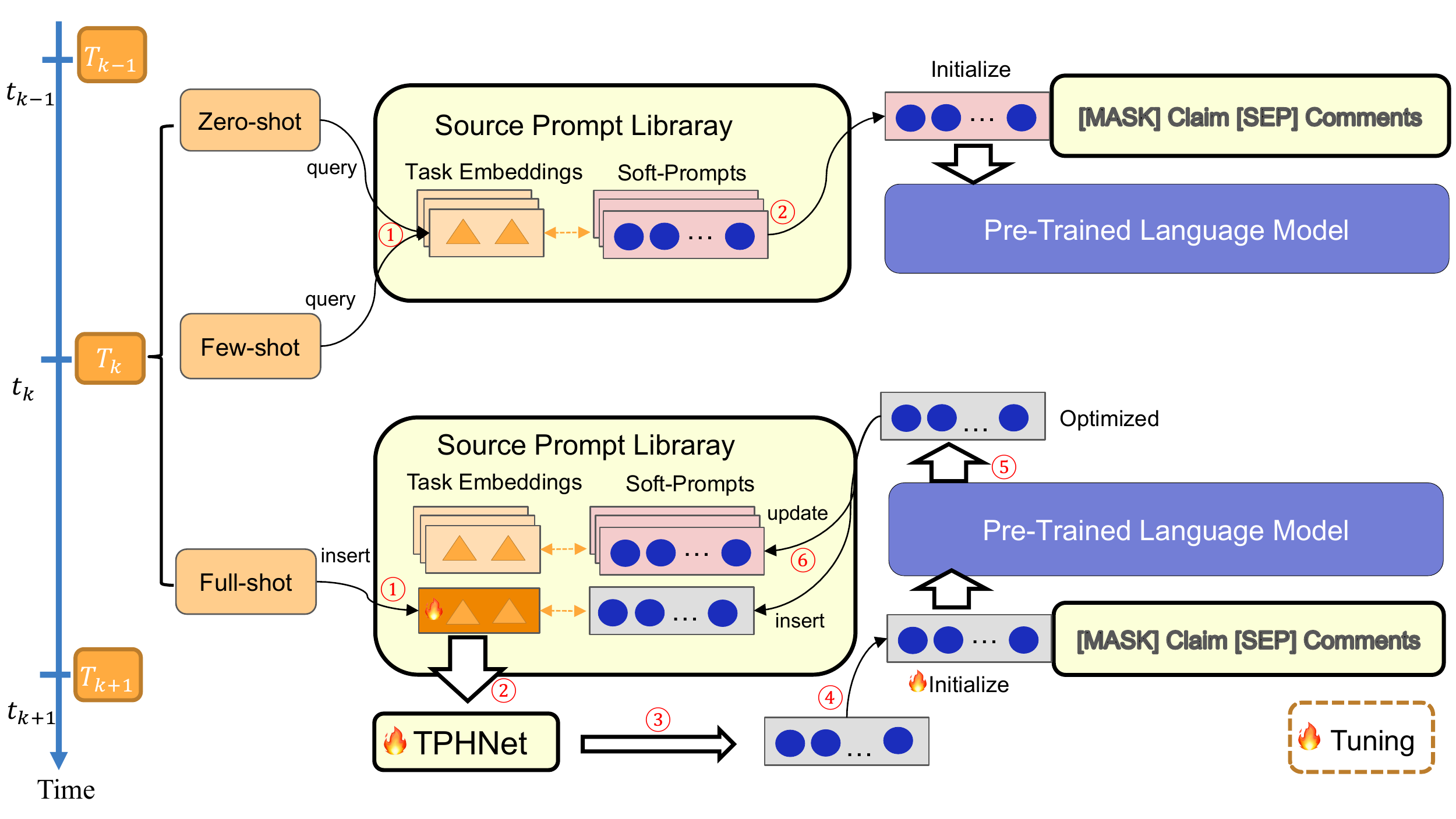}
		\vspace{-0.3cm}
		\caption{Illustration of the execution of CPT-RD in a single task in CDET.}
		\label{fig:model}
		\vspace{-0.3cm}
	\end{figure*}
	
	\section{Methodology}
	\label{sec:method}
	In this section, we first present the task definition and execution process of CPT-RD in CDET, then elaborate prompt encoding method and bidirectional knowledge transfer strategy. 
	
	\subsection{Task definition and description}
	Suppose RD model $\mathcal{M}_{1:k-1}$ has performed learning on a sequence of tasks from $1$ to $k-1$, denoted as $\mathcal{T}_{1:k-1} =\{\mathcal{T}_1,\ldots,\mathcal{T}_{k-1}\}$. Each task is a domain-specific rumor binary classification problem (non-rumor or rumor), and the inputs are claim and comments texts.
	The goal of the RD model is to use the knowledge gained from upstream $\{\mathcal{T}_{1:k-1}\}$ tasks to help learn a better detector $\mathcal{M}_{1:k}$ for the $k$-th task $\mathcal{T}_{k}$ while avoiding forgetting knowledge learned from past tasks. We use the terms “domain” and “task” interchangeably because each task is from a different domain. 
	
	Often, each task consists of three stages: zero-shot stage, few-shot stage, and full-shot (large-scale labeled training) stage, which corresponds to the development process of rumors from germination to normalization as described in Sec.\ref{sec:intro}. 
	As shown in Fig.\ref{fig:model}, in the zero-shot and few-shot stages, CPT-RD applies various forward knowledge transfer strategies proposed in Sec.\ref{sec:fwt} to query the soft-prompt as the prompt initialization of the current task. Note that the soft-prompt of the current task is randomly initialized if the source prompt library (SPL) is empty.
	In the full-shot phase of the current task, CPT-RD starts training with a randomly initialized soft-prompt instead of using the soft-prompt in SPL because, in experiments, we found that the former works better. CPT-RD obtains knowledge accumulation in full-shot annotated samples through TPHNet. The optimized soft-prompt specific to the current task is stored in the SPL at the end of full-shot training.

	\subsection{Prompt encoding}
	
	Prompt tuning (PT) formalizes RD as a masked language modeling problem using a pre-trained language model (PLM).
	In Fig.\ref{fig:ptuning}, given the $k$-th task input, the pretrained embedding layer $e$ of PLM converts the claim and comments text into token embeddings $\mathbf{X}_k \!\!=\! e(\text{claim}) \in \mathbb{R}^{m\times d}$ and $\mathbf{C}_k \!\!= \!e(\text{comments}) \in \mathbb{ R}^{n\times d}$, $m,n$ is the token length, and $d$ is the hidden dimension of PLM.
	We prepend $l$ randomly initialized soft-prompt tokens $\mathbf{P}_k =\{P_{k}^1,P_{k}^2,\ldots P_{k}^l\} \in \mathbb{R}^{l\times d}$ before them, where $P_{k}^i\in \mathbb{R}^d$ is an embedding vector. Then, we add a $\mathtt{[MASK]}$ token, which is used to predict the label words $y\in \mathcal{Y}$. The input embedding of PLM is:
	\begin{equation}
		\label{eq:input_seq}
		\mathbf{x} = \mathbf{P}_k,\mathtt{[MASK]},\mathbf{X}_k, \mathtt{[SEP]}, \mathbf{C}_k,
	\end{equation}
	and only $\mathbf{P}_k$ is learnable.
	Hence the tuned parameters in PT are extremely fewer than full-parameter fine-tuning, which is friendly for model deployment.

	One key ingredient of PT is the \emph{verbalizer}: a mapping from the class label to a word token in the PLM vocabulary. 
	PLM gives the probability of each word $v$ in the vocabulary being filled in $\mathtt{[MASK]}$ token $p(\mathtt{[MASK]}=v|\mathbf{x})$. To map the probabilities of words into the probabilities of labels, we define the \emph{verbalizer} as $ver$, which form the label word set $\mathcal{V}$, to the label space $\mathcal{Y}$, i.e., $ver\colon \mathcal{V} \mapsto \mathcal{Y}$. We use $\mathcal{V}_y$ to denote the subset of $\mathcal{V}$ that is mapped into a specific label $y$,  $\cup_{y\in\mathcal{Y}} \mathcal{V}_y = \mathcal{V}$. Then the probability of label $y$, i.e., $p(y|\mathbf{x})$, is calculated as:
	\begin{equation}
		p(y|\mathbf{x}) \!=\! f_{ver}\left(p(\mathtt{[MASK]}\!=\!v|\mathbf{x})|v\in\mathcal{V}_y\right),
		\vspace{-1mm}
	\end{equation}
	where $f_{ver}$ is a function transforming the probability of label words into the probability of the label. 
	The training objective is:
	\begin{equation}
		\label{eq:loss1}
		L_{{\mathbf{P}_k}} = -\!\!\!\sum_{\mathbf{x}_i,y_i \in D_{\mathcal{T}_k}}\!\! \log(p(y_i|\mathbf{x}_i)),
	\end{equation}
	where $D_{\mathcal{T}_k}$ is defined as the training data for task $\mathcal{T}_k$.
	
	PT only inserts soft-prompt into the input embedding sequence of PLM. 
	We follow the deep prompt tuning of P-tuning v2, adding the prompts of different layers as prefix tokens to the input sequence independently of other layers to increase tunable task-specific parameters and improve stability.
	In experiment Sec.\ref{sec:main}, CPT-RD evaluates the above two soft-prompt placement methods respectively.

	\subsection{Forward knowledge transfer}
	\label{sec:fwt}
	Due to the lack of samples or even no training samples (zero-shot and few-shot) in the early stage of rumor events, it is necessary to forward transfer the knowledge from upstream tasks for fast detection. An intuitive way of knowledge transfer is to reuse knowledge gained from previous tasks, which often improves and accelerates learning for future tasks. 
	
	Therefore, SPL is responsible for storing upstream soft-prompts $\{\mathbf{P}_j\}_{j<k}$ and task embeddings $\{\mathbf{z}_j\}_{j<k}$, where $\mathbf{P}_j$ is obtained after training on $D_{\mathcal{T}_k}$ in the full-shot stage, $\mathbf{z}_j=\frac{1}{|D_{\mathcal{T}_k} |} \sum_{i=1}^{|D_{\mathcal{T}_k}|} f_e(\mathbf{X}^i_k, \mathbf{C}^i_k)$, $f_e$ is an encoder model, i.e. BERT.
	Based on SPL, three types of prompt initialization are proposed.
	
	\textbf{CLInit:} Use the previous task's soft-prompt $\mathbf{P}_{k-1}$ to initialize the current task's soft prompt $\mathbf{P}_k$. 
	\textbf{SimInit:} Select $\mathbf{P}_k$ from $\{\mathbf{P}_j\}_{j<k}$ with the highest similarity to the current task representation $\mathbf{z}_k\in \mathbb{R}^d$ for initialization. Note that $\mathbf{z}_k$ is calculated based on the training data of task $\mathcal{T}_k$, which does not need to be labeled.
	Straightforwardly, we compute euclidean distances $E$ and cosine similarities $C$ for task embedding pairs in the two groups and use the averaged results as the final similarity metrics:
	\begin{eqnarray}
		\begin{aligned}
			\mathrm{E}(\mathbf{z}_{k},\mathbf{z}_{j}) &=\frac{1}{1+\Vert{\mathbf{z}}_{k}-{\mathbf{z}}_{j}\Vert}, \\
			\mathrm{C}(\mathbf{z}_{k},\mathbf{z}_{j}) &=\frac{{\mathbf{z}}_{k}\cdot{\mathbf{z}}_{j}}{\Vert{\mathbf{z}}_{k}\Vert\Vert{\mathbf{z}}_{j}\Vert}.
		\end{aligned}
	\end{eqnarray}
	When CPT-RD faces the upstream task domain samples again, it directly reuses the soft-prompt of the corresponding task, thus completely avoiding the occurrence of CF, but CLInit and SimInit are single-source transfer strategies. 
	
	Single-source reuse initialization strategy can completely avoid CF, but only a single task is considered in the forward knowledge transfer. Intuitively, knowledge available for transfer should be present in all upstream tasks. 
	\textbf{MeanInit:} Calculate the average of $\{\mathbf{P}^u_j\}_{j<k}$ to obtain $\mathbf{P}_k$. In deep prompt tuning, each layer of soft-prompt is correspondingly averaged. MeanInit considers multi-source transfer, but none of the above strategies can accumulate to transfer backward knowledge. We empirically compare these three strategies in Sec.\ref{sec:main}.

	\subsection{TPHNet for backward knowledge transfer}
	\label{sec:bwt}
	Although storing the training-optimized $\mathbf{P}_k$ after labeled train on task $\mathcal{T}_k$ can avoid forgetting, it ignores the backward knowledge transfer of the tasks.
	Hypernetwork is usually used to consolidate the knowledge of sequence tasks and has a certain ability of backward knowledge transfer \cite{von2019continual,kj2020meta,jin2021learn}. It is a network that explores the meta-parameter space of another network.
	Similar to Hypernetwork, instead of modeling the final classification result, task-conditioned prompt-wise hypernetwork (TPHNet) learns a latent distribution space of soft-prompt with task-specific priors, aiming to accumulate knowledge in sequential and use it for all future tasks.
	
	Specifically, when the CPT-RD completed the learning of task $\mathcal{T}_{k-1}$, and before the full-shot phase of $\mathcal{T}_k$ starts, the current task embedding $\mathbf{z}_k$ and the soft-prompt set $\{\mathbf{P}_j\}_{j<k}$ already exist in SPL as described in Sec.\ref{sec:fwt}.
	Here, the task representation $\mathbf{z}_k$ for task $\mathcal{T}_k$ is optimized jointly while learning the task. 
	Taking the PT case as an example, adding soft-prompt to the embedding layer, deep prompt tuning only needs to modify the generation dimension of soft-prompt. 
	TPHNet $g$ generates a soft-prompt $\mathbf{P}_k$ through an auto-encoder, using $\mathbf{z}_k$ as input:
	\begin{equation}
		g(\mathbf{z}_k) = W_2(\text{tanh}(W_1 \mathbf{z}_k +b_1)) +b_2,
	\end{equation}
	where $W_1\in \mathbb{R}^{d'\times d}, b_1\in \mathbb{R}^{d'}, W_2\in \mathbb{R}^{(l\times d) \times d'}$ and $b_2\in \mathbb{R}^{l \times d'}$ are trainable parameters, $d'=64$ is the middle dimension. 
	
	Then, in each step of learning $\mathcal{T}^i_k$, we randomly sample a prior task soft-prompt $\mathbf{P}_j (j < k)$ to regularize the TPHNet learning. It penalizes the $\ell_2$ distance between the soft-prompt generated at the current step ${\mathbf{P}^{i}_k} = g(\mathbf{z}^i_k)$ and the pre-computed one, \textit{i.e.}, $\lvert \lvert \mathbf{P}^{i}_k - {\mathbf{P}}_{j} \rvert \rvert_2^2$.
	Therefore, we avoid the TPHNet changes its output for a prior task too much during the sequential task learning, so that the knowledge accumulation is better guaranteed for the learned model. The following overall loss function:
	\begin{equation}
		L_{\mathcal{T}_k} = \hat{L_{{\mathbf{P}_k}}} + \frac{\beta}{k-1}\sum_{i=1}^{k-1} \lvert \lvert \mathbf{P}^{i}_k - {\mathbf{P}}_{j} \rvert \rvert_2^2 ,
	\end{equation}
	where $\hat{L_{{\mathbf{P}_k}}}$ is the cross entropy like Eq.\ref{eq:loss1}. According to this equation, optimizing the overall loss will update the patameters of the TPHNet $g$, task embedding $\mathbf{z}_{k}$ and prompt-based PLM.
	$\beta=0.01$ is a hyperparameter that controls the strength of the regularizer. After training, if $\mathbf{P}_{k}$ obtains better (or the same) performance than $\mathbf{P}_j$ on $\mathcal{T}_j$, we update $\mathbf{P}_j$ to $\mathbf{P}_{k}$.
	%



	\begin{table*}[ht]
		\centering
		
		\begin{adjustbox}{scale=0.7,center}
			\begin{tabular}{l|cccc|cccc|cc}
				\toprule
				& \multicolumn{4}{c|}{PHEME + Twitter15\&16 + Covid19} & \multicolumn{4}{c|}{Weibo }  & \multicolumn{2}{c}{} \\
				\midrule
				Method &  Avg.F1  & FWT  & BWT   & fs.F1(val) & Avg.F1 & FWT  & BWT  & fs.F1(val) &+Params & Tune Params\\
				\midrule
				Fine-tuning$^\dagger$  &   54.6  ($\pm$ 1.2) &  7.0  & -24.3 & 67.5& 45.6 ($\pm$3.1) & 18.3     & -23.2 &75.6 & 0& 100\% \\
				EANN$^\dagger$  &   57.9  ($\pm$ 1.5) &  8.1  & -20.3 & 68.3& 49.4 ($\pm$2.1) & 18.5     & -20.9 &77.1 & 1.8\% & 100\% \\
				Adapter$^{\star}$       &  66.7($\pm$ 2.5) & 13.7 & -14.2 & 74.6 &58.0 ($\pm$2.3) & 24.4   & -17.5  &83.9 & 2.1\% &  2.1\%\\
				ParallelAdapter$^{\star}$       & 61.5 ($\pm$ 2.2) & 2.8 &   \best -4.7 & 68.7 & 55.8 ($\pm$2.5) &   8.5 & -7.1 &75.3 & 2.3\% &  2.3\%\\
				prompt-tuning (CLS)$^\star$       & 60.0 ($\pm$ 1.5) & 7.4    & -15.7& 75.9 & 67.8 ($\pm$1.8) &  18.3  & -5.1 & \best 84.1 & 0.03\% & 0.03\%\\
				p-tuning v2 (CLS)$^{\star}$         & 62.9  ($\pm$ 2.1) & 5.1      & -9.3  & 73.0  & 67.6 ($\pm$1.6) &  24.4 & \best -3.0 & 80.4&  0.6\%&  0.6\%\\
				prompt-tuning (VER)$^\star$       & 65.0 ($\pm$ 1.5) & 7.1    & -10.2 & 75.7   & 70.5 ($\pm$1.8) &  18.7  & -7.5   & 82.8 & 0.03\% & 0.03\%\\
				p-tuning v2 (VER)$^{\star}$         &\best  69.7  ($\pm$ 1.1) & \best 13.9   & -11.0  &\best 76.3  & \best 71.8  ($\pm$1.6) &  \best 29.5 & -6.8 & 82.6&  0.6\%&  0.6\%\\
				\midrule
				\multicolumn{11}{c}{Prompt-tuning based (PT-based)}  \\
				\midrule
				CPT-RD (CLInit)$^{\star}$   & 65.1 ($\pm$ 0.8) & 15.0 & 0 & 75.5 & 70.0 ($\pm$0.7) & 18.3 & 0  & 79.0&  0.03\%&  0.03\% \\
				CPT-RD (SimInit)$^{\star}$     & 66.3($\pm$ 0.9) &\best 15.8 & 0 &\best 76.7 &  71.6($\pm$1.6) & 19.8  & 0   &83.6  &0.03\%&  0.03\%\\	
				CPT-RD (MeanInit)$^{\star}$     & 64.5($\pm$ 1.2) & 14.8 & 0 & 75.6 & 70.3($\pm$1.5) & 18.5  & 0   & 80.9 &0.03\%&  0.03\%\\		
				CPT-RD (SimInit+TPHNet)$^{\star}$     &\best 66.7($\pm$ 1.3) & 15.3 &\best 0.2 & 76.0  &\best 71.9($\pm$1.2) &\best 20.0  &\best 0.5   &\best 84.0  & 0.1\%&  0.1\%\\
				\midrule
				\multicolumn{11}{c}{P-tuning v2 based (PTv2-based)}  \\
				\midrule
				CPT-RD (CLInit)$^{\star}$   & 68.2 ($\pm$ 1.0) & 19.8 & 0  &76.1 & 73.1($\pm$0.7) & 30.7 & 0  & 83.2&  0.6\%&   0.6\% \\
				CPT-RD (SimInit)$^{\star}$     & 72.2($\pm$ 1.1) & 23.3 & 0 & 79.1 & 75.0($\pm$1.2) & 31.2  & 0  & 84.9 &  0.6\%&  0.6\%\\
				CPT-RD (MeanInit)$^{\star}$     & 70.6($\pm$ 1.5) & 20.5 & 0 & 75.5 & 73.5($\pm$1.3) & 28.7  & 0  & 83.0 &  0.6\%&  0.6\%\\
				CPT-RD (SimInit+TPHNet)$^{\star}$     &\best 75.0($\pm$ 1.3) &\best 23.5 &\best 0.9 &\best 79.3&\best 76.5($\pm$1.6) &\best 31.4  &\best 1.1  &\best 85.7 &  1\%&  1\%\\
				\bottomrule
			\end{tabular}
			
		\end{adjustbox}
		\caption{\label{tab:results} Results evaluated on test datasets for all tasks in \texttt{PHEME + Twitter15\&16 + Covid19} and \texttt{Weibo}. The following averaged over 5 random task orders (Table \ref{tab:task_order}) are reported, where ($\star$) and ($\dagger$) indicate frozen language model parameters and fine-tuning, (+Params) and (Tune Params) are additional parameters and the tunable parameters for each task. CLS and VER denote the output predictions with CLS token classifier and Verbalizer head, respectively. The fs.F1 is the result on the validation dataset, and the k-shot is 16. }
		\vspace{-0.5cm}
	\end{table*}

	\begin{figure*}[tb]
		\centering
		\subfigure[k=16 during sequential tasks]{
			\begin{minipage}[t]{0.48\linewidth}
				\centering
				\includegraphics[width=\linewidth]{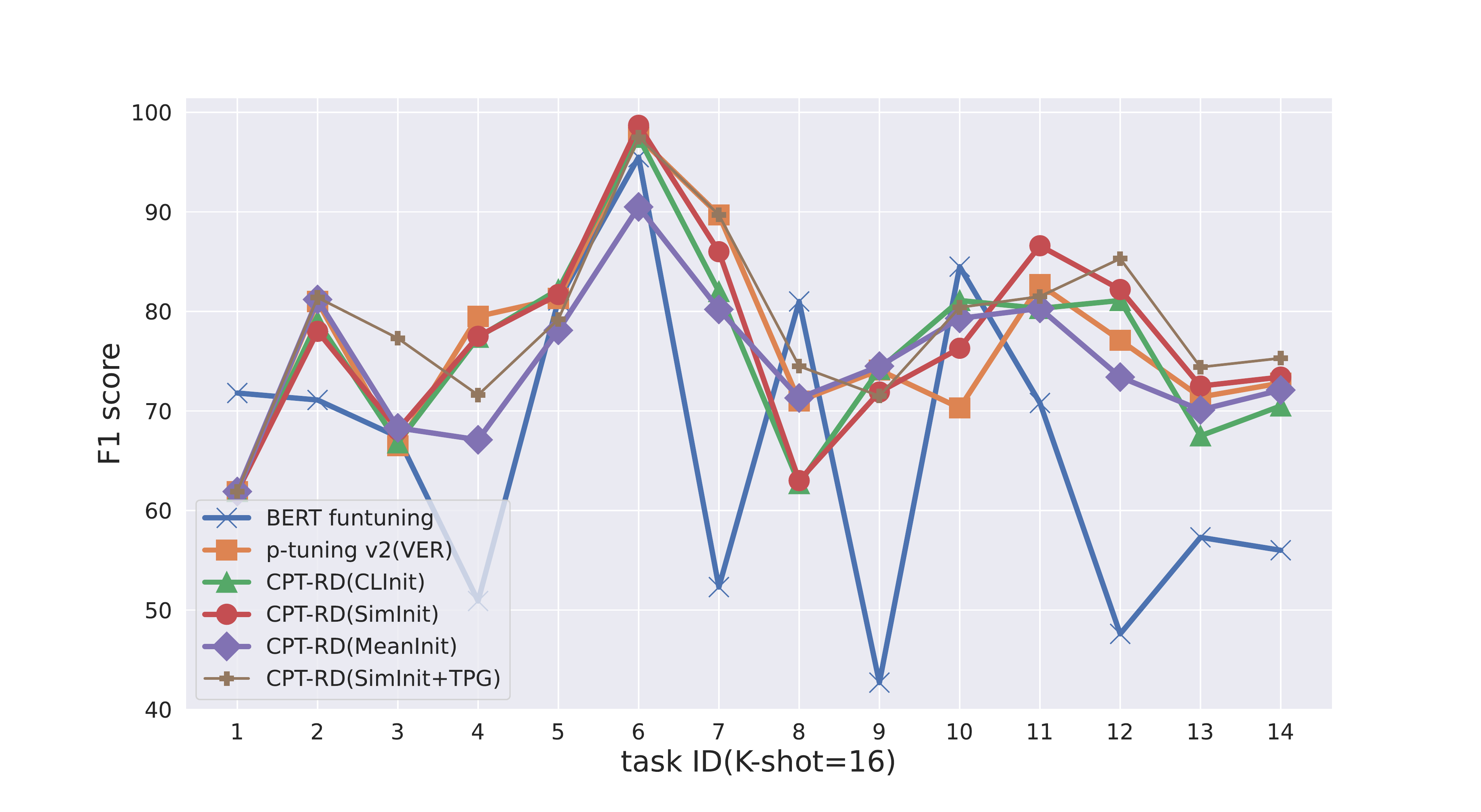}
			\end{minipage}
		}
		\subfigure[The average fs.F1 performance for k=4,8,16]{
			\begin{minipage}[t]{0.48\linewidth}
				\centering
				\includegraphics[width=\linewidth]{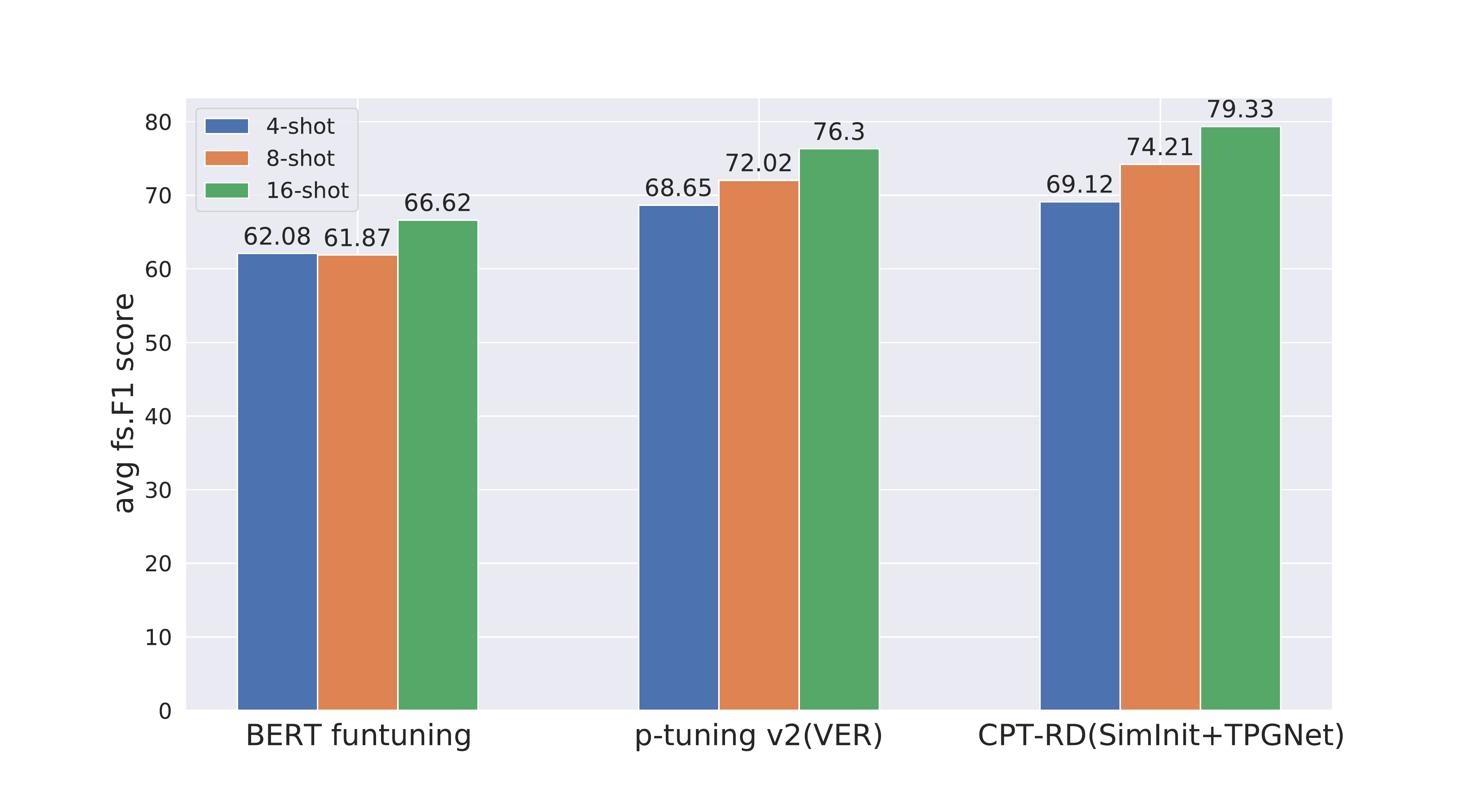}
			\end{minipage}
		}
		\vspace{-0.3cm}
		\caption{The few-shot performance of each model on the dataset \texttt{PHEME + Twitter15\&16 + Covid19}.}
		\label{fig:fewshot}
		\vspace{-0.5cm}
	\end{figure*}

	\section{Experiments}
	\label{sec:analysis}
	
	To evaluate the proposed CPT-RD, we closely follow the settings proposed in prior works \cite{lopez2017gradient,jin2021learn}, and conduct comprehensive experiments. In particular, we mainly consider whether CPT-RD effectively addresses the four challenges mentioned in Sec.\ref{sec:intro}. We carefully compare CPT-RD with state-of-the-art methods of different categories under proper experiment settings. Moreover, we conduct extensive ablation studies to provide a deeper understanding of our method.

	\subsection{Datasets and evaluation metrics}
	\label{sec:datasets}
	We collected 14 domain events for each dataset, Chinese and English. 
	Each piece of data contains a claim, comments, and label (non-rumor or rumor). We split the data for each event into train/validation/test datasets with a split ratio of 30\%/35\%/35\%. The details about the statistics are shown in Table \ref{table:data_statistics}, and the PHEME and Weibo \cite{lu2021novel} datasets are well divided by event domain. We also added the COVID-19 dataset \cite{patwa2021overview} for the English dataset, which comes from the competition\footnote{https://competitions.codalab.org/competitions/26655\href{https://competitions.codalab.org/competitions/26655}.}. In addition, for datasets without split of events, including Twitter15 and Twitter16 \cite{ma2018detect}, we use Tweeter-LDA \cite{diao2012finding}, an LDA variant widely used for short and noisy tweets, to determine topic clusters as well as important words with their weights. We removed the label of unverified rumors in Twitter15 and Twitter16, retained the rumors and non-rumor, and divided them into five-event domains. The similarity between the datasets we calculated using TF-IDF is shown in Fig.\ref{fig:correlation}. To prepare the model inputs for both datasets, first, we replace the URLs with the special token “$\mathtt{[unused10]}$".  Then, we also replace the usernames with the special token “$\mathtt{[unused11]}$". 
	These two datasets will be released on GitHub along with our experimental code. 
	
	To evaluate the performance of RD using F1 score as a metric, $R_{j,i}$ is defined as the F1 score on the test set of task $\mathcal{T}_i$ after training on task $\mathcal{T}_j$.
	We follow the two indicators, FWT and BWT, proposed by \citet{lopez2017gradient} to evaluate the knowledge transfer ability of CPT-RD in the process of continual learning. 
	We evaluate the average F1 performance for all tasks after full-shot training on the final task $\mathcal{T}_N$:
	\begin{equation}
		\mbox{\normalsize Avg.F1}  =  \frac{1}{N}
		\sum_{i=1}^N R_{T,i} . \label{eq:acc} 
	\end{equation}
	where $N$ is the number of tasks. According to \citet{lopez2017gradient}, two metrics are defined to measure the effect of forward and backward transfers:
	\begin{eqnarray}
		\mbox{ \normalsize BWT} & = & \frac{1}{N-1}
		\sum_{i=1}^{N-1} R_{N,i} - R_{i,i}   \label{eq:bwt} ,\\
		\mbox{\normalsize FWT} & = & \frac{1}{N-1}
		\sum_{i=2}^{N} R_{i-1,i} - R_{0,i}. \label{eq:fwt} 
	\end{eqnarray}
	FWT is the average zero-shot performance on a new task and evaluating the model's generalization ability. BWT assesses the impact of learning on the subsequent task has on the previous task. A negative BWT indicates that the model has forgotten some previously acquired knowledge.

	In addition, to evaluate the model’s performance early in the birth of the rumor, we also recorded the average few-shot performance of each new task during the continual cumulative training, which is a cumulative value:
	\begin{equation}
		\mbox{\normalsize fs.F1}  =  \frac{1}{N}
		\sum_{i=1}^N R^{fs}_{i,i}, \label{eq:fs}
	\end{equation}
	where $R^{fs}_{i,i}$ denotes the F1 performance of training on the few-shot training set of the $i$-th task and testing on the test set of the $i$-th task.

	\subsection{Compare models}
	\label{sec:baseline}
	In the experiments, BERT-base is used as the PLM weights, and CPT-RD will be compared with the following advanced models: \textbf{Fine-tuning:} Fine-tune the model on new task data continually. \textbf{EANN:} \citet{wang2018eann} uses adversarial networks to extract event invariant features for generalization when new events arrive. For comparison, EANN is modified to use BERT as the encoder and only text data. \textbf{Adapter:} Freeze the pre-trained model and train a residual Adapter\cite{houlsby2019parameter}. \textbf{ParallelAdapter:} A variant by transferring the parallel insertion of prefix tuning into adapters \cite{he2021towards}. \textbf{prompt-tuning (CLS/VER):} which only tunes soft-prompts with a frozen language model \cite{lester2021power}, prompt for transformer's first layer. \textbf{p-tuning v2 (CLS/VER):} Using multilayer soft-prompts (deep prompt tuning), where CLS/VER denotes the predicted output with $\mathtt{[CLS]}$ token and Linear classification layer and verbalizer MLMhead, respectively. \textbf{MTL p-tuning v2 (CLS/VER):} P-tuning v2 in a multi-task manner instead of CL. Train a single prompt using all tasks’ data concurrently.

	\subsection{Main results}
	\label{sec:main}
	The higher the FWT score, the better the model works for the unseen domain. A negative BWT indicates that the model produces forgetting, and if it is positive, it can accumulate knowledge.
	
	\begin{table*}[htbp]
		\centering
				\begin{adjustbox}{scale=0.64,center}
					\begin{tabular}{cccccccccc}
						\toprule
						Task ID & Task           & Fine-tuning & p-tuning v2(VER)   & CPT-RD(CLInit)  & CPT-RD(SimInit) & CPT-RD(MeanInit)  & CPT-RD(SimInit+TPHNet) \\
						\midrule
						1     & Charlie Hebdo              & 22.1  & 64.1  & 75.3                     & 77.1                   &75.3                 & 78.0   \\
						2     & TwitterEvent4             & 65.1  & 88.0   &\secbest  72.1    & \secbest65.2      &\secbest70.6   &\secbest74.4      \\
						3     & Ferguson                      & 24.7  & 62.4  &  70.3                   &  72.7                  &67.5                 & 72.8   \\
						4     & Germanwings-crash &  50.6  & 63.4 & 65.9                    &\secbest60.6        &64.7                 &\secbest61.3    \\
						5     & Ottawa Shooting        &  52.7  & 74.3   &\secbest 70.5   & \secbest72.2        &74.6                 & 78.3   \\
						6     & Prince Toronto            & 98.7  &  71.4   &  98.1                   &   95.6                 &93.7                 &\secbest94.3    \\
						7     & Putin missing               &  53.0  &  55.3 &\secbest53.0   & 66.0                      &\secbest53.5  & 72.1    \\
						8     & TwitterEvent1              &  77.1  &  77.1  &   77.1                   &    77.8                 &\secbest70.7   &\secbest74.5   \\
						9     & Sydney Siege               & 42.8  &  72.5   &  74.2                  &\secbest65.7     &75.8                  &74.9    \\
						10    & TwitterEvent5             &  71.4  & 67.5   & 69.8                     &     70.5             &71.7                   &73.5   \\
						11    & Gurlitt                            &  43.7  & 48.3   &\secbest42.2   &   62.3                  &52.5                  &62.5   \\
						12    & Covid19                        &  47.6  & 86.7   & 87.9                    &\secbest84.3   &\secbest85.3     &88.3     \\
						13    & TwitterEvent2            &   57.8  & 69.0  &\secbest60.5     &  70.8                   &\secbest65.2     &72.4    \\
						14    & TwitterEvent3            &  56.9  & 74.9   &\secbest 37.9    &\secbest70.9       &\secbest67.3     &\secbest71.3    \\
						\midrule
						& Avg.                                      & 54.6  & 69.7   &\secbest68.2  & 72.2      &  70.6   & 75.0    \\
						\bottomrule
			\end{tabular}
		\end{adjustbox}
		\caption{\label{tab:table2}The F1 score evaluated on the final model after all 14 tasks are visited in the test set. We use Avg. to represent the average F1 of all tasks for each method. The gray numbers indicates that CPT-RD (p-tuning v2 based) does not perform as well as normal p-tuning v2 in a single task.}  
		\vspace{-0.62cm}
	\end{table*}%
	
	\textbf{Compare model performance.}
	In Table \ref{tab:results}, unsurprisingly, the Fine-tuning model has a severely CF, the BWT value of -24.3 under the CL setting. The level of catastrophic forgetting in EANN \cite{wang2018eann} is somewhat reduced compared to Fine-tuning but is still severe. Prompt-tuning and p-tuning v2 are somewhat related to the adapter method in the form of parameter tuning \cite{he2021towards}, but their performance in CL differs. The prompt-based model is better than the adapter in both datasets. 
	From Tabel \ref{table:multi-task} in Sec.\ref{sec:mtl}, consistent with \citet{liu2021p}, CLS token classification is better than the verbalizer when p-tuning v2 is under multi-task learning mode. The performance of multi-task learning is usually considered the upper limit of the CL model. However, verbalizer's FWT and final F1 scores in the CL setting were better than CLS but more prone to CF. Since SPL lets the model not generate CF, we next develop CPT-RD using prompt-tuning (PT) and p-tuningv2 (PTv2) baseline with verbalizer.

	\textbf{CPT-RD performance.} 
	We have developed CPT-RD based on PT and PTv2 respectively. We can find that since CPT-RD has SPL, the BWT values are all greater than or equal to 0, which indicates that this fundamentally eliminates the CF. The more stable performance among the three forward transfer strategies is SimInit. CLInit is the most unstable, probably because the soft-prompt from the previous task is not necessarily beneficial to the learning of the next task and is prone to cumulative negative effects on subsequent tasks in CL. 
	It can be observed by Tabel \ref{tab:table2} that the F1 score of CLInit on the last two tasks of the last sequence task is significantly lower than other strategies, 60.5 and 37.9 respectively.
	MeanInit with multi-source prompt does not work as expected and is only slightly better than CLInit. Finally, we added TPHNet to SimInit to achieve backward knowledge transfer, but the performance on PT is not stable enough, probably because PT has fewer tunable parameters. Furthermore, TPHNet incorporates additional tunable parameters but still has parameter-efficient that is much lower than Fine-tuning.

	\textbf{Final model performance.} 
	The final model performance is the model's performance after learning the last task in the sequence.
	In Table \ref{tab:table2}, although CPT-RD solves CF, in the ideal case, the final model performance should outperform PTv2. We can find that the final performance of CLinit is inferior to PTv2 in 7 tasks, simInit and meanInit in 6, and SimInit+TPG in 5, which further indicates that TPHNet achieves a certain degree of backward knowledge transfer. Among them, we can find that in CLinit, the last task is as low as 37.9, indicating that directly reusing the soft-prompt of the previous task may have a more significant impact on the subsequent tasks.
	
	\textbf{Data privacy and Parameter explosion.}
	From Sec.\ref{sec:bwt}, we propose TPHNet in the context of hypernetwork, which avoids rehearsal-based data replay and thus preserves data privacy. For a more comprehensive evaluation of our method, we replace TPHNet with a rehearsal-based technique, more details in Sec.\ref{sec:rehearsal}. From Table \ref{table:rehearsal}, it can be found that TPHNet can achieve comparable performance to the rehearsal-based method.
	
	Besides, it can be seen from Sec.\ref{sec:fwt} that CPT-RD avoids parameter explosion by using SPL to store soft-prompt (without dynamic model expansion). And the storage space occupied by SPL is negligible.
	In Table \ref{tab:table2}, the tuning parameter of the Adapter-based model is 2.1-2.3\%, which is more parameter-efficient than the 100\% of the Fine-tuning model. However, CPT-RD adds only 0.03-1\% tunable parameters based on more parameter-efficient PT and can achieve better performance.
	
	\textbf{Few-shot performance.} 
	There is a lack of samples in the early burst stages, so few-shot performance is critical for RD models. The final model is the model obtained after learning the last task. From Fig.\ref{fig:fewshot} (a), it can be found that Fine-tuning has a very unstable few-shot performance in CL. PTv2 still shows better few-shot capability than Fine-tuning in CL, which indicates that the PT-based model is more suitable for domain generalization. In Fig.\ref{fig:fewshot} (b), the average fs.F1 of CPT-RD (SimInit+TPHNet) is higher than that of PTv2, indicating that continuously accumulated knowledge can be better used for few-shot.

	 \subsection{CPT-RD with rehearsal buffer}
	\label{sec:rehearsal}
	To verify the gap between TPHNet and rehearsal-based technology, we randomly sample 50 pieces of data from each domain training set in memory, and jointly trained on the future domain dataset.
	It can be observed from Table \ref{table:rehearsal} that the rehearsal-based technology reflects the certain ability to backward knowledge transfer by accumulating data in past domain tasks. 
	From Table \ref{tab:results}, the BWT value of TPHNet is 0.9, which shows that our method is close to rehearsal-based technology, but our advantage is that the memory occupies small, and there is no data privacy problem.
	\begin{table}[h]
		\centering
		\small
		\resizebox{\linewidth}{!}{
			\begin{tabular}{llccc}
				\toprule
				\textbf{Buffer size}& \textbf{Methods (PTv2-based)} &  \textbf{Avg. F1}  & \textbf{FWT} & \textbf{BWT} \\ \midrule
				\multirow{4}{*}{0/domain}&p-tuning v2 (VER) & 69.7 & 13.9  & -11.0 \\
				&CPT-RD (CLInit) &   68.2 &  19.8  & 0 \\
				&CPT-RD (MeanInit) & 70.6 & 20.5  & 0 \\
				&CPT-RD (SimInit) &  72.2 & 23.3  & 0 \\
				\midrule
				\multirow{4}{*}{50/domain}&p-tuning v2 (VER) & 71.2 & 14.1  & 1.5 \\
				&CPT-RD (CLInit) & 72.1 & 20.1  & 1.1 \\
				&CPT-RD (MeanInit) & 71.7 & 19.4  & 0.8 \\
				&CPT-RD (SimInit) & 75.3 & 23.8  & 1.2\\
				\bottomrule
			\end{tabular}
		}
		\caption{The performance of CPT-RD (PTv2-based) on the \texttt{PHEME + Twitter15\&16 + Covid19} (English) dataset using rehearsal-based technology.}
		\label{table:rehearsal}
		\vspace{-0.5cm}
	\end{table}

	\subsection{Multi-task learning performance}
	\label{sec:mtl}
	The performance of multi-task learning methods is often defined as an upper bound on the performance of continual learning. Multi-task learning avoids CF by visiting data from different domain tasks at different times. In Table \ref{table:multi-task}, we experimented with the model's performance in the multi-task learning mode under the Chinese and English datasets. 
	Consistent with the experimental conclusion of \citet{liu2021p}, p-tuning v2 using CLS token classification outperforms the verbalizer under multi-task learning.
	However, the verbalizer outperforms CLS in FWT and final F1 score in the CL setting, but is more prone to CF.
	\begin{table}[h]
		\centering
		\small
		\resizebox{\linewidth}{!}{
			\begin{tabular}{llccc}
				\toprule
				\textbf{Methods} &  \textbf{Avg. F1}  & \textbf{+Params} & \textbf{Tune Params} \\ \midrule
				\multicolumn{4}{c}{ PHEME + Twitter15\&16 + Covid19 }  \\
				\midrule
				MTL p-tuning v2 (CLS) &   80.3 ($\pm$ 0.8)  &  0.6\%  & 0.6\% \\
				MTL p-tuning v2 (VER) &   77.9 ($\pm$ 0.8)  & 0.6\% & 0.6\% \\
				\midrule
				\multicolumn{4}{c}{ Weibo }  \\
				\midrule
				MTL p-tuning v2 (CLS) &   85.3 ($\pm$ 0.8)  &  0.6\%  & 0.6\% \\
				MTL p-tuning v2 (VER) &   83.8 ($\pm$ 0.8)  & 0.6\% & 0.6\% \\
				\bottomrule
			\end{tabular}
		}
		\caption{ The performance of the baseline model p-tuning v2 in the multi-task learning mode. }
		\label{table:multi-task}
		\vspace{-0cm}
	\end{table}
		
	\subsection{Ablation study}
	To understand the effectiveness of the different techniques proposed, we conducted an ablation study. From Table \ref{tab:results}, we know that CPT-RD is more stable in PTv2-based than PT-based, which indicates that the model can benefit from more tunable parameters. Conventional PT do not use forward transfer strategy and TPHNet with lower FWT values. In PTv2-based, the overall performance is highest when SimInit and TPHNet are used together. In Table \ref{table:xiaorong}, removing either of TPHNet or SimInit will result in a decrease in overall performance. This shows the validity of our improvements, which would degrade to p-tuning v2 if SimInit and TPHNet were both removed. 

	\begin{table}[!t]
		\begin{center} 
			\scalebox{0.9}
			{
				{
					\setlength\tabcolsep{0.5em}
					\begin{tabular}{l|ccc}
						\toprule
						{\textbf{model}} & \multicolumn{1}{c}{ \textbf{Avg.F1}} & \multicolumn{1}{c}{ \textbf{FWT}} & \multicolumn{1}{c}{ \textbf{BWT}} \\
						\midrule
						{CPT-RD} & \multicolumn{1}{c}{75.0} & \multicolumn{1}{c}{\enspace 23.5} &    \multicolumn{1}{c}{\enspace0.9}\\  
						{w/o TPHNet} & \multicolumn{1}{c}{72.2} & \multicolumn{1}{c}{\enspace 23.3} & \multicolumn{1}{c}{\enspace0}\\ 
						{w/o SimInit} & \multicolumn{1}{c}{71.3} & \multicolumn{1}{c}{\enspace14.2} &    \multicolumn{1}{c}{\enspace0.9}\\  
						{w/o SimInit+TPHNet} & \multicolumn{1}{c}{69.7} & \multicolumn{1}{c}{\enspace13.9} & \multicolumn{1}{c}{\enspace-11.0}\\ 				
						\bottomrule
			\end{tabular}}}
			\caption{Ablation study on the effectiveness of the CPT-RD (SimInit+TPHNet) PTv2-based on the test set in the \texttt{PHEME + Twitter15\&16 + Covid19}. }
			\label{table:xiaorong}
			\vspace{-0.7cm}
		\end{center} 
	\end{table}

	\section{Conclusion}
	\label{sec:conclusion}
	We explore how to continually detect a social network environment with frequent unseen domains in RD. 
	The novel framework Continual Prompt-tuning RD (CPT-RD) is proposed, which includes various knowledge transfer techniques. 
	In the face of emergency domains, CPT-RD can use soft-prompt initialization strategies to achieve fast generalization. There is also a task-conditioned prompt-wise generator network (TPHNet) in terms of continual accumulation of detection knowledge. 
	Future works include extending usage of CPT-RD to task agnostic scenarios and designing more diverse knowledge transfer strategies.
	
	\bibliography{sample-base}

	\clearpage
	\appendix
	\section{Appendix}
	


	\subsection{Event correlation}
	\label{sec:correlations}
\begin{figure*}[tb]
		\centering
		\subfigure[\texttt{PHEME + Twitter15\&16 + Covid19}]{
			\begin{minipage}[t]{0.48\linewidth}
				\centering
				\includegraphics[width=\linewidth]{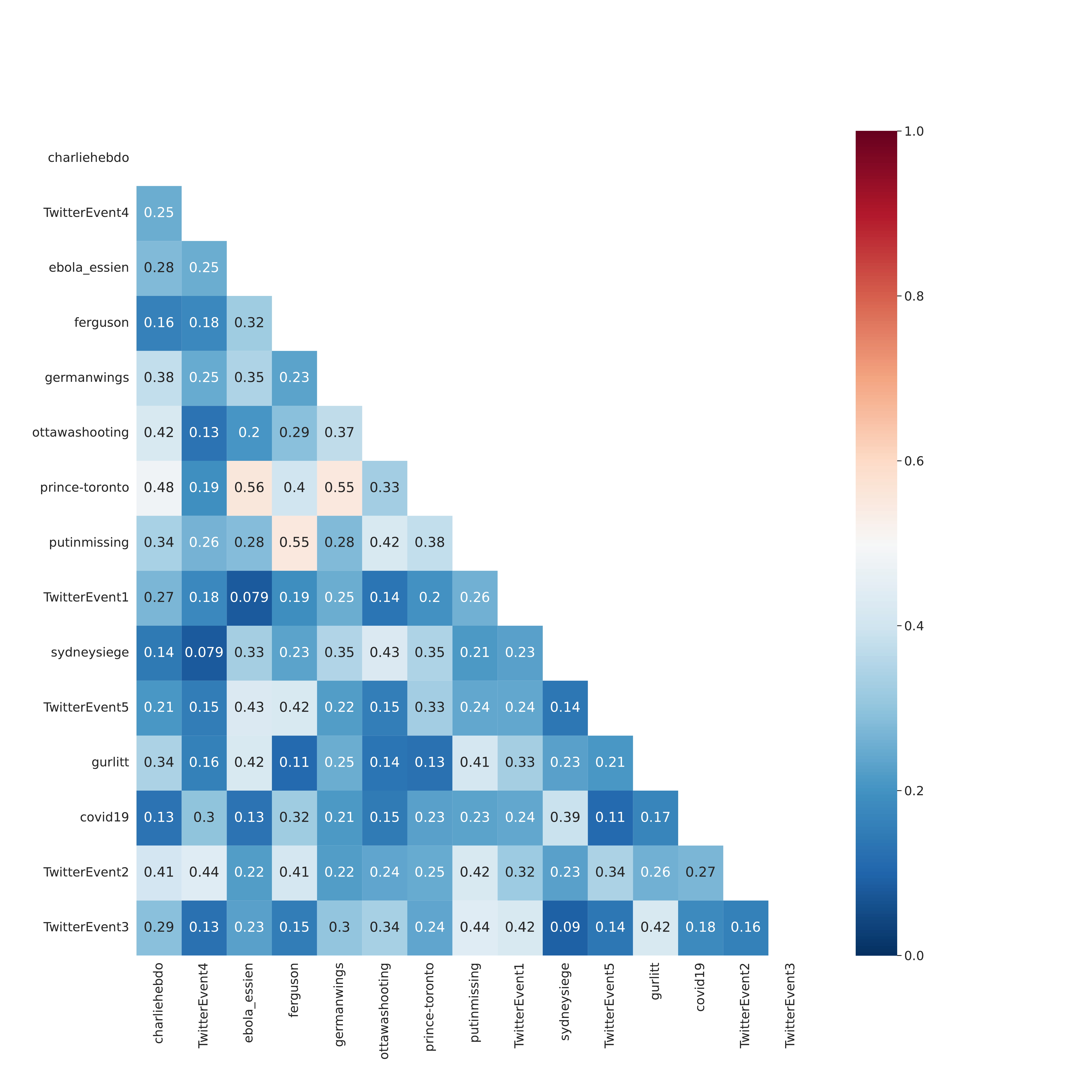}
			\end{minipage}
		}
		\subfigure[\texttt{Weibo}]{
			\begin{minipage}[t]{0.48\linewidth}
				\centering
				\includegraphics[width=\linewidth]{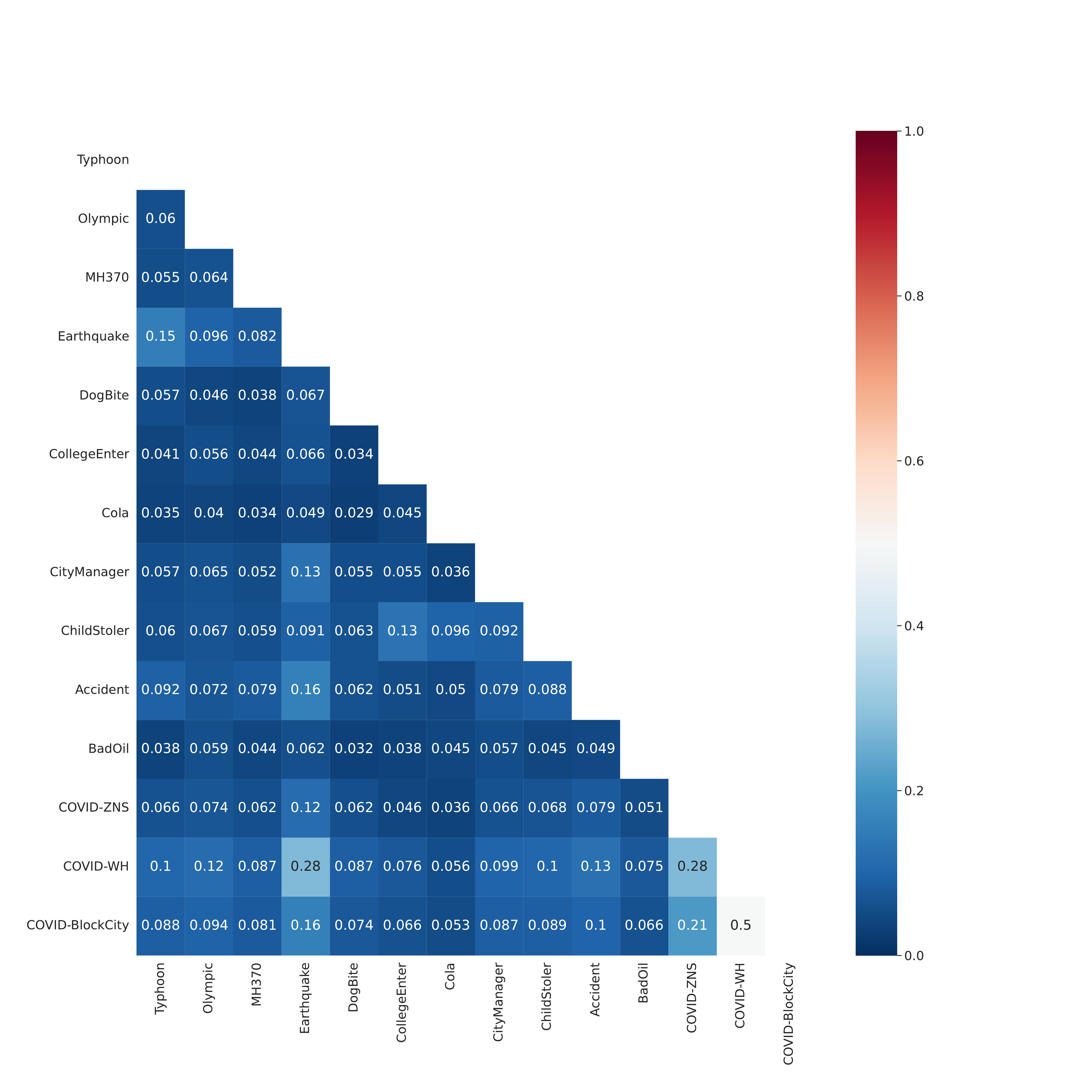}
			\end{minipage}
		}
		\vspace{-0.3cm}
		\caption{Domain Task Relevance Heat Map.}
		\label{fig:correlation}
		\vspace{-0.5cm}
	\end{figure*}

	After filtering the stopping words, we use TF-IDF to calculate the event relevance of each task domain, as shown in Fig.\ref*{fig:correlation}.

	\begin{table}[htbp]
		\centering
		\small
		\resizebox{\linewidth}{!}{
			\begin{tabular}{llccc}
				\toprule
				\textbf{Datasets}& \textbf{tasks} &  \textbf{non-rumor}  & \textbf{rumor} & \textbf{total} \\ \midrule
				&Gurlitt & 77 & 61  & 138 \\
				&Putin missing & 112 & 126  & 238 \\
				&Prince Toronto & 4 & 229  & 233 \\
				PHEME&Germanwings-crash & 231 & 238  & 469 \\
				&Ferguson & 859 & 284 & 1143  \\
				&Charlie Hebdo & 1621 & 458  & 2079 \\
				&Ottawa Shooting & 420 & 470  & 890 \\
				&Sydney Siege & 699 & 522  & 1221 \\ 
				&TwitterEvent 1 & 63 & 156  & 219 \\ 
				&TwitterEvent 2 & 89 & 164  & 253 \\ 
				Twitter15\&16&TwitterEvent 3 & 153 & 198  & 351 \\ 
				&TwitterEvent 4 & 146 & 200  & 346 \\ 		
				&TwitterEvent 5 & 128 & 436  & 564 \\ 
				Covid19&Covid19 & 3060 & 3360  & 6420 \\ 
				\midrule
				&total & 7662 & 6902  & 14564 \\
				\bottomrule
				&MH370 & 133 & 262  & 395 \\
				&Olympics & 173 & 81  & 254 \\
				&Urban managers & 94 & 149  & 243 \\
				&Cola & 215 & 419  & 634 \\
				&Child trafficking & 94 & 172  & 266 \\
				&Waste oil & 133 & 57  & 190 \\
				Weibo&Accident & 100 & 82  & 182 \\
				&Earthquake & 117 & 58  & 175 \\
				&Typhoon & 107 & 64  & 171 \\
				&Rabies & 101 & 42  & 143 \\
				&College entrance exams & 147 & 590  & 737 \\ 
				&Lockdown the city & 86 & 24  & 110 \\
				&Zhong Nanshan & 55 & 21  & 76 \\
				&Wuhan & 167 & 69  & 236 \\
				\midrule
				&total & 1722 & 2048  & 3770 \\ 
				\bottomrule
			\end{tabular}
		}
		\caption{Statistics of \texttt{PHEME + Twitter15\&16 + Covid19} (English) and \texttt{Weibo} (Chinese) datasets.}
		\label{table:data_statistics}
		\vspace{-0.5cm}
	\end{table}

	\subsection{Implementation details}
	\label{sec:details}
	We tune hyperparameters on the \texttt{PHEME + Twitter15\&16 + Covid19} and \texttt{Weibo} validation sets. We tune learning rates by enumerating over [3e-3, 5e-3, 7e-3], and finally use a learning rate of 7e-3 for all CPT-RD approaches where a learning rate of 1e-4 for TPHNet. The learning rate of the fine-tuning approaches is 5e-5, and the learning rate of the adapter approaches is 1e-4. Regarding the PT and PTv2, including the MTL approach, we use a learning rate of 5e-3. We use a batch size of 16 across experiments. We train the model for at most 100 epochs for each training task with a patience of 4 epochs without validation performance improvement. Before training on a new task, we revert the model to the checkpoint with the best validation performance in the previous task. In the few-shot learning stage, we use the same learning rate and train the model for 500 steps ($k \in \{16, 8, 4\}$), assuming no validation sets to perform early stopping. The length of the input sequence on data set \texttt{PHEME + Twitter15\&16 + Covid19} is 300, and \texttt{Weibo} is 128. We also experimented with the length of different soft-prompt tokens, such as $\{20, 40, 60, 80\}$, and found that the performance of 40/60 is relatively stable, so 40 is uniformly used as the length of soft-prompt in the experiment.

	\begin{table*}[htbp]
	\centering
	\scalebox{0.7}{
		\begin{tabular}{ lp{19cm}c }
			\toprule
			Task Order & \multicolumn{1}{c}{Tasks} \\ \midrule
			\multicolumn{2}{c}{\textbf{\texttt{PHEME + Twitter15\&16 + Covid19}} } \\
			Order1 &  Charlie Hebdo, TwitterEvent4, Ferguson, Germanwings-crash, Ottawa Shooting, Prince Toronto, Putin missing, \\ &TwitterEvent1, Sydney Siege, TwitterEvent5, Gurlitt, Covid19, TwitterEvent2, TwitterEvent3    \\
			Order2 &  Sydney Siege, ferguson,TwitterEvent1, Gurlitt, Ottawa Shooting, TwitterEvent3, Prince Toronto, TwitterEvent4,\\ & Putin missing, Charlie Hebdo, TwitterEvent5, Germanwings-crash, TwitterEvent2, Covid19    \\
			Order3  & TwitterEvent3, Charlie Hebdo, TwitterEvent2, Ferguson, TwitterEvent1, TwitterEvent5, TwitterEvent4, Putin missing, \\ &Ottawa Shooting, Prince Toronto,Gurlitt, Germanwings-crash, sydneysiege, Covid19    \\
			Order4  &  TwitterEvent1, Germanwings-crash, TwitterEvent4, Ferguson, Gurlitt, Sydney Siege, TwitterEvent3, Charlie Hebdo, \\ &Ottawa Shooting, Prince Toronto,TwitterEvent2, TwitterEvent5, Putin missing, Covid19    \\
			Order5 &   Covid19, Germanwings-crash, Prince Toronto, TwitterEvent3, TwitterEvent5, Sydney Siege, Ferguson, Ottawa Shooting, \\ &Charlie Hebdo,TwitterEvent4, TwitterEvent2, Gurlitt, Putin missing, TwitterEvent1     \\
			\midrule
			\multicolumn{2}{c}{\textbf{\texttt{Weibo}} }   \\
			Order1   &   Typhoon, Olympic, MH370, Earthquake, Rabies, College entrance exams, Cola, Urban managers, Child trafficking, Accident, Waste oil, Zhong Nanshan, Wuhan, Lockdown the city \\
			Order2  &    Olympic, Child trafficking, Rabies, Accident, Earthquake, Cola, College entrance exams,Typhoon, Urban managers, Waste oil, MH370, Zhong Nanshan, Wuhan, Lockdown the city \\
			Order3  &    Zhong Nanshan, Wuhan, Lockdown the city, Cola, Accident, Urban managers, Waste oil, Earthquake, Olympic, Child trafficking,Typhoon, Rabies, College entrance exams, MH370 \\
			Order4  &    Cola, Accident, Olympic, Waste oil, Typhoon, Zhong Nanshan, Wuhan, Lockdown the city, College entrance exams, Urban managers, Rabies, Child trafficking, MH370, Earthquake \\
			Order5   &    Zhong Nanshan, Wuhan, Lockdown the city, MH370, Urban managers, Child trafficking, Typhoon, Earthquake, Olympic, Cola, Accident, Rabies, Waste oil, College entrance exams  \\
			\bottomrule
		\end{tabular}
	}
	\caption{Order of continual learning tasks in \texttt{PHEME + Twitter15\&16 + Covid19} and \texttt{Weibo} datasets.}
	\label{tab:task_order}
\end{table*}

\end{document}